# GuidelineGuard: An Agentic Framework for Medical Note Evaluation with Guideline Adherence

MD Ragib Shahriyear
Augmedix Bangladesh
Dhaka, Bangladesh

*Abstract*—Although rapid advancements in Large Language Models (LLMs) are facilitating the integration of artificial intelligence-based applications and services in healthcare, limited research has focused on the systematic evaluation of medical notes for guideline adherence. This paper introduces GuidelineGuard, an agentic framework powered by LLMs that autonomously analyzes medical notes, such as hospital discharge and office visit notes, to ensure compliance with established healthcare guidelines. By identifying deviations from recommended practices and providing evidence-based suggestions, GuidelineGuard helps clinicians adhere to the latest standards from organizations like the WHO and CDC. This framework offers a novel approach to improving documentation quality and reducing clinical errors.

## I. INTRODUCTION

Healthcare guidelines play a crucial role in providing evidence-based care, protecting patients from medical malpractice, and improving clinical outcomes. Adherence to these guidelines reduces variations and risks in patient management, decreases risks of medical malpractice, as well as improves the quality of treatment [1]. Strict guideline adherence also improves the efficiency of healthcare systems, as they help streamline treatment processes and resource utilization in hospitals, clinics, and emergency care units. Non-compliance with guidelines in critical conditions, such as heart or kidney failure, and seizures can result in increased hospitalizations, complications, and even wrongful death due to medical negligence [2]. Clinical studies highlighted that poor adherence to guidelines, for example, sepsis management, has resulted in delayed treatments, hence worsening patient outcomes [3]. These examples clearly show that guidelines contribute a lot towards reducing mistakes made by healthcare practitioners.

While developing healthcare applications, Large Language Models (LLMs) present a major challenge due to their inherent knowledge cut-offs, as they are trained on datasets that do not continuously update with the latest information. This limitation can be particularly problematic in domains that require real-time access to the most up-to-date knowledge, such as healthcare guidelines. As healthcare guidelines frequently get updated based on new clinical evidence and research findings, general LLMs do not have access to the most up-to-date information, potentially limiting their reliability when applied to medical decision-making [4].

Healthcare guidelines are often dispersed across various sources, some of which may not be authoritative or trustworthy. This could lead to the LLM relying on outdated or incorrect guidelines, further worsening the potential for clinical errors [5]. A more robust and flexible solution to this challenge involves using standardized sets of healthcare guidelines embedded directly into the LLM framework. By employing guideline embeddings—where the guidelines themselves are transformed into vectors that the model can reference—LLMs can evaluate medical notes against a predefined and reliable set of criteria. This method offers the advantage of transparency and control, as the system explicitly knows which guidelines are being used to assess the medical notes. Furthermore, this approach allows for flexibility; if the clinical context or healthcare system changes, the system can be updated to reference a different set of guidelines without retraining the entire model.

Embedding specific guidelines, such as those from the World Health Organization (WHO) or the Centers for Disease Control and Prevention (CDC), ensures that medical notes are evaluated based on verified, authoritative criteria. This can help reduce clinical errors and malpractice risks by aligning clinical documentation and decision-making with evidence-based standards [6]. Embedding standard healthcare guidelines within LLM systems offers a more controlled and reliable method for evaluating medical notes, thus mitigating the risk of relying on potentially outdated or inaccurate information. This approach provides a scalable solution for maintaining adherence to healthcare guidelines in dynamic clinical environments.

Despite significant ongoing research into the application of large language models (LLMs) in healthcare, limited attention has been given to the evaluation of medical notes for adherence to established guidelines, particularly through the combination of advanced techniques such as Retrieval-Augmented Generation (RAG) and agentic workflows. This project addresses this gap by demonstrating the potential of LLMs in the automated evaluation of clinical documents

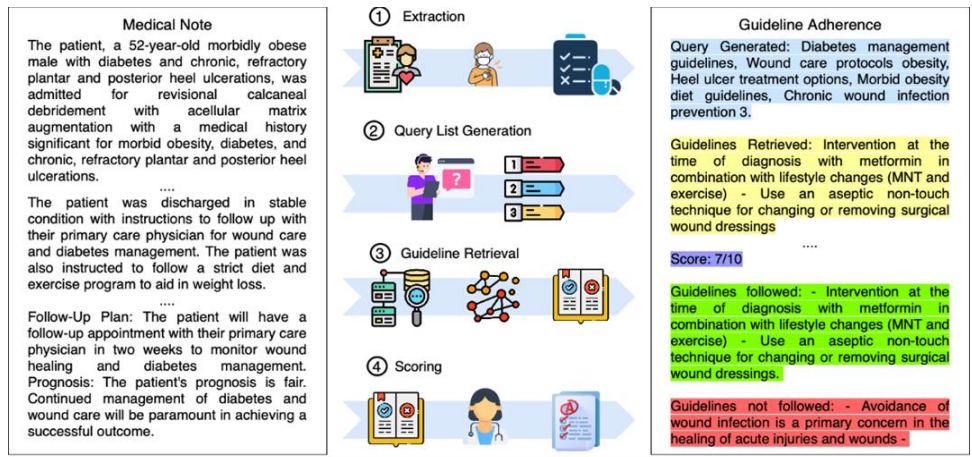

Fig. 1. Demonstration of GuidelineGuard

against healthcare guidelines, yielding promising results. To the best of our knowledge, this is the first study to focus specifically on the use of LLMs for guideline-based evaluation of medical notes. In this project, we are presenting the following work:

We present GuidelineGuard, an agentic framework that evaluates medical notes based on healthcare guideline adherence, with best practices and results of our experiments. The code, prompts, and all relevant resources will be open-sourced after publications for the community to use, recreate, and improve the work.

We aspire to contribute meaningfully through our work in the advancement of healthcare and the ongoing development of research in Large Language Model based applications.

## II. RELATED RESEARCH

With the rapid advancements in large language models (LLMs), an increasing number of studies are investigating their potential applications in healthcare. These research works cover various use cases, such as medical note generation, note correction, diagnostic support, decision-making, information extraction, and the integration of LLMs with chatbots for patient interaction and personalized healthcare plans.

### A. Medical Note Generation

LLMs are expected to play a pivotal role in medical note generation. By leveraging state-of-the-art automatic speech recognition (ASR) and natural language processing (NLP), LLMs can transcribe patient-clinician conversations and generate medical notes with remarkable accuracy. Results in this area have been promising, suggesting that such models could streamline documentation tasks significantly [7]. Researchers are also conducting comparative studies to evaluate the performance of general-purpose models, such as GPT-4, against specialized medical LLMs [8]. Ongoing investigations focus on generating specialty-specific notes and hospital discharge summaries to enhance clinical efficiency, with encouraging results emerging from these initiatives [9].

### B. Medical Note Correction

Beyond generating medical notes, LLMs exhibit substantial potential in note correction. This application extends beyond the correction of automatically generated notes, as LLMs can also rectify errors in clinician-authored notes, thereby improving the accuracy and quality of clinical documentation. A promising approach involves using Named Entity Recognition (NER) to identify disease-related terms, followed by employing Retrieval-Augmented Generation (RAG) for contextual understanding derived from external datasets. LLMs then act as engines for refining and correcting the clinical notes, which results in achieving higher documentation standards for healthcare [10].

### C. Diagnosis and Medical Decision-Making

The convergence of LLMs and visual language models holds significant promise in clinical decision-making. Models such as Vision Transformers and state-of-the-art LLMs are delivering impressive results in tasks related to diagnostic support [11]. Healthcare practitioners can leverage LLMs for a range of tasks, including conducting research, enhancing reasoning capabilities, and facilitating question-answering activities [12]. Moreover, LLMs are proving effective in providing personalized recommendations for treatment plans, thereby enhancing patient care through customized interventions [13], [14].

### D. Retrieval-Augmented Generation (RAG) and Healthcare Applications

RAG has emerged as a powerful technique for LLM applications, as it enables developers and researchers to access the most recent information while maintaining confidentiality in data-driven decision-making. RAG has shown promising results, particularly in the generation of clinical trial documentation, further demonstrating its utility across diverse healthcare contexts [15]. Additionally, leading LLMs such as those in the GPT series, Google's Gemini, and Claude-3-Opus, are demonstrating significant potential when integrated with RAG techniques [16]. In addition to document generation, RAG is proving to be a highly effective tool for improving question-answering systems, enhancing their accuracy and relevance in medical contexts [17], [18].

## III. GUIDELINEGUARD SETUP

To evaluate the guideline adherence of medical notes using LLMs, there are several components to set up. First, the guidelines on which the medical notes will be evaluated. Second, to fetch the correct guidelines for a given medical note, we did some processing of the guideline dataset and turned it into an embedding with all guideline documents. Third, we made an agentic framework and distributed the task of guideline adherence evaluation among 4 agents. Each agent has been given a different task, which they complete utilizing the tool and LLM given to them. The LLM we chose in the work is the open-source Llama-3 70B by Meta [19]. The details of the setup can be found in the following section.

*A. Healthcare Guidelines*

Healthcare guidelines help providers enhance patient care quality, meet regulatory standards, and achieve better patient outcomes [20]. We used the healthcare guideline dataset published by the EPFL LLM Team on HuggingFace, which was also used to train the MediTron-70B model [21]. This dataset contains 37K articles extracted from sources that allow content redistribution, namely CCO, CDC, CMA, ICRC, NICE, PubMed, SPOR, WHO, and WikiDoc.

*B. Document Embeddings*

Embeddings have become a pivotal means to represent complex, multi-faceted information about entities, concepts, and relationships in a condensed and useful format [22]. We used the pubmedbert-base-embeddings-matryoshka embedding from HuggingFace, which is a version of PubMedBERT Embeddings with Matryoshka Representation Learning applied. Matryoshka embeddings are a method of representing nested structures in data using embeddings that encapsulate multiple layers of information [23]. We used an embedding size of 768 to convert the guideline chunks into embeddings. For retrieving the documents, we used cosine similarity. Although in 'cosine similarity' larger values indicate closer proximity, unlike smaller values in distance metrics, it has also become a widely used measure of semantic similarity between entities, since directional alignment of learned embedding vectors is considered more important than the magnitude of the vectors [24].

*C. Agentinc Framework*

We used LangGraph, a popular library by LangChain to build the agentic framework. [25]. We designed 4 agents: Extractor Agent, Query Agent, Retriever Agent, and Scorer Agent, each with their specific prompt and tools.

- Extractor Agent: The Extractor Agent is responsible for extracting the diagnoses and treatments mentioned in a given medical note. It is crucial to identify the diagnoses and treatments, as they will be used later to score the note in terms of guideline adherence.

- Query Agent: The Query Agent analyzes the medical note and generates several queries. The purpose of these generated queries is to retrieve healthcare guidelines that are relevant to the note. The number of queries generated by this agent can be experimental. For simplicity and keeping things fast, we kept the number of queries = 5, but we have also experimented with 7 queries, and also generated queries for every diagnosis mentioned in the note.

- Retriever Agent: The Retriever Agent retrieves relevant healthcare guidelines using the queries generated by the query agent. These guidelines are paramount for measuring the guideline adherence of a medical note. For every query, the retriever agent retrieves several guideline documents.

- Scorer Agent: The scorer agent gives the medical note a score based on guideline adherence. For every diagnosis mentioned in the note, if the treatments mentioned follow the guidelines properly, the Scorer Agent will give the note a positive score of +1. If the note includes a treatment that does not follow the guidelines, the Scorer Agent will give the note a negative score of +1. Furthermore, if the note mentions any diagnosis, but does not mention any treatment for that said diagnosis, the Scorer Agent will give a negative score of +1 as well. After the Scorer Agent gives scores for all diagnoses and treatments, a final score will be generated.

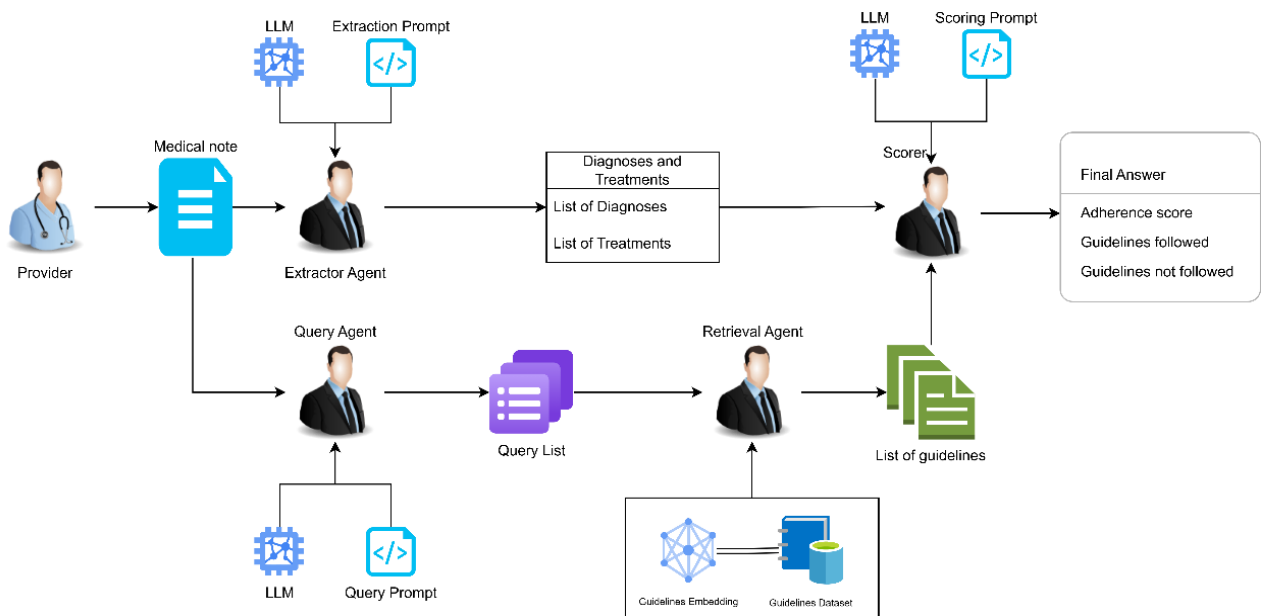

Fig. 2. Overview of the Agentic Framework

## IV. EXPERIMENT AND RESULTS

To evaluate the GuidelineGuard framework, we have experimented with multiple medical notes from different specialties. Each note was evaluated with a score between 0 and 1, 0 being no guidelines followed and 1 being all guidelines followed. The length of the medical notes range from 300-1000 words. The average results of the experiment are shown in Table 1.

TABLE I. STATISTICS OF GUIDELINEGUARD FRAMEWORK EXPERIMENT WITH MEDICAL NOTES FROM MULTIPLE SPECIALTIES

| Specialty | Followed | Not followed | Score |
|---|---|---|---|
| Family Medicine | 1.5 | 0.5 | 0.75 |
| Pediatrics | 1.0 | 1.0 | 0.50 |
| Cardiology | 1.0 | 0.5 | 0.67 |
| Oncology | 1.0 | 0.5 | 0.67 |
| Endocrinology | 2.0 | 0.5 | 0.80 |
| Pulmonology | 1.5 | 2.0 | 0.43 |
| Orthopedics | 1.0 | 1.0 | 0.50 |
| Gastroenterology | 2.5 | 1.0 | 0.17 |

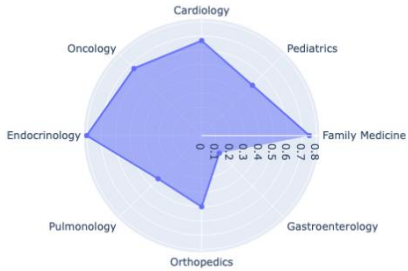

Fig. 3. Average guideline adherence score of medical notes from different specialties

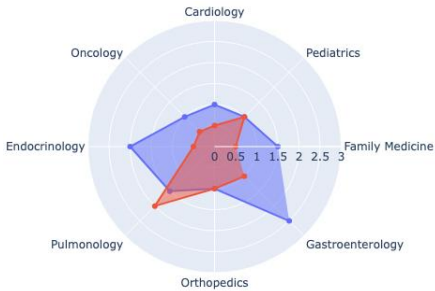

Fig. 4. Average number of guidelines followed (blue) and not followed (red) for calculating guideline adherence score.

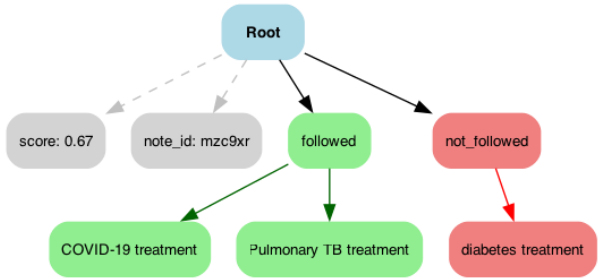

Fig. 5. Graphviz view of the guideline adherence of medical note for better visualization.

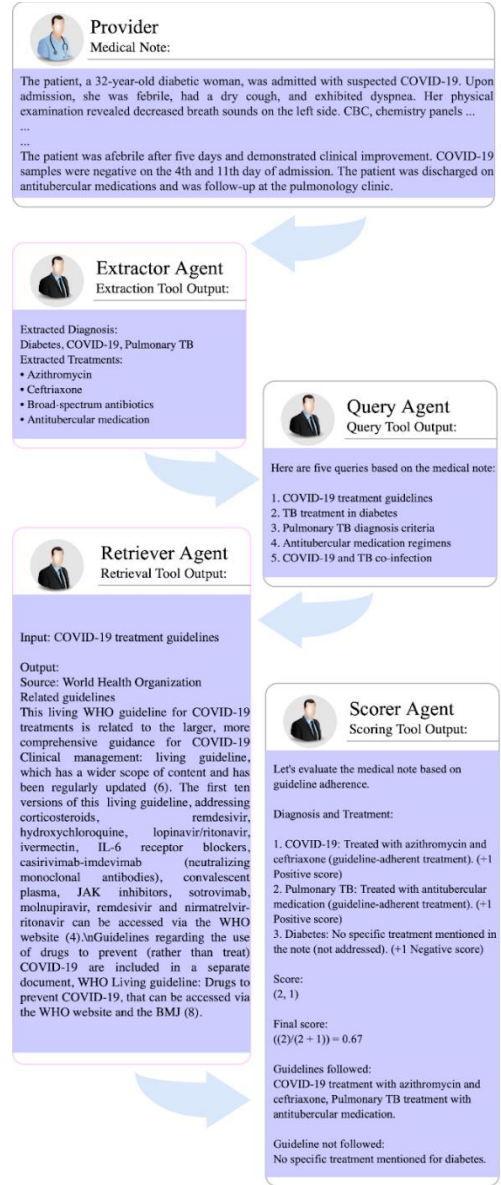

Fig. 6. Workflow and tool output of multiple agents to evaluate a note based on guidelines.

## V. CONCLUSION

In this paper, we presented GuidelineGuard, an agentic framework for evaluating medical notes based on guideline adherence. GuidelineGuard consists of extracting, query, retrieval, and scoring modules. Extensive evaluations demonstrate that GuidelineGuard is capable of rigorous evaluation of medical notes from all specialties based on guidelines adherence, underscoring the untapped potential of LLMs in healthcare and offering promising avenues for the intersection of AI and healthcare. GuidelineGuard can be used in multiple scenarios, from improving patient care to analyzing health systems, as well as directing upcoming research on healthcare guidelines.

## VI. FUTURE WORK

This study offers valuable insights into medical note evaluation and improving patient care in health systems. In the future, we would like to compare the performance of other LLMs using the framework, such as GPT-4o, Llama 3.1 405B, Claude 3 Opus, and Mixtral 8x22B [19], [26], [27],

[28]. We plan on continuing the research further by working with multimodal models and incorporating image data, such as x-rays, EKG signals, and MRI images into the framework and analyzing them using agents for guideline adherence, as multimodal models are continuing to get better. Another aspect of our future work is evaluating healthcare systems using the data generated by the GuidelineGuard framework such as note scores, list of guidelines followed, etc.

## VII. Limitations and Ethical Concerns

This study offers valuable insights into medical note evaluation and improving patient care in health systems, but there are a few limitations to consider.

First, because LLMs are trained on vast amounts of text data, they may inadvertently capture and reproduce biases present in the data. For instance, they might favor particular recommendations concerning a given diagnostic or associate certain medical disorders with particular demographics. As a result, the agents might give false information and favor one set of guidelines over another. Furthermore, important details that are necessary to assess whether a treatment is appropriate or not are absent from the notes used to evaluate the framework. These details include the numerical results of lab tests, imaging data, and other patient information. Agents may therefore overanalyze some diagnoses or have hallucinations.

Second, considering the privacy concerns, we limited the test of our proposed framework with synthetic medical notes, since LLMs can present privacy hazards if used real medical notes, potentially violating HIPAA regulations.

When developing AI systems for healthcare, it is crucial to rigorously consider biases, privacy concerns, and safety implications. Ensuring these factors are addressed is essential for creating ethical and reliable AI technologies in medical contexts.